\documentclass{edm_template}

\usepackage{booktabs} 
\usepackage{amsmath,amssymb,amsfonts}
\usepackage{algorithmic}
\usepackage{graphicx}
\usepackage{textcomp}
\usepackage{xcolor}
\usepackage{tabularx}

\usepackage[T1]{fontenc}

\usepackage{soul}
\usepackage[utf8]{inputenc}
\usepackage{caption}
\usepackage{multirow}
\usepackage{multicol}
\usepackage{graphicx}
\usepackage{bm}
\usepackage{threeparttable}
\usepackage{tabularx}
\usepackage{tabu}
\usepackage{amssymb}
\usepackage{adjustbox}
\usepackage{booktabs}
\usepackage{array}
\usepackage{subcaption}
\usepackage[title]{appendix}
\usepackage{url}

\begin{document}

\title{Academic Performance Estimation with Attention-based Graph Convolutional Networks}
%
%
%
%
%

\numberofauthors{2} 
%

%
%
\author{
\alignauthor
Qian Hu\\
       \affaddr{Department of Computer Science}\\
       \affaddr{George Mason University}\\
       \affaddr{Fairfax, Virginia}\\
       \email{qhu3@gmu.edu}
\alignauthor
Huzefa Rangwala\\
       \affaddr{Department of Computer Science}\\
       \affaddr{George Mason University}\\
       \affaddr{Fairfax, Virginia}\\
       \email{rangwala@cs.gmu.edu}
}

\maketitle
\begin{abstract}

%
Student's academic performance prediction empowers 
educational technologies 
including academic trajectory and degree planning, 
course recommender systems, early warning and 
advising systems. Given a student's past 
data (such as grades in prior courses),
the task of student's performance prediction is to predict a student's grades in future courses.
Academic programs are structured in a way that prior courses lay the foundation for future courses. 
The knowledge required by courses is obtained by taking multiple prior courses, which exhibits complex 
relationships modeled by graph  structures. Traditional 
methods for student's performance prediction usually 
neglect the underlying relationships between multiple courses;
and how students acquire knowledge across them. In addition, traditional 
methods do not provide interpretation for predictions needed for 
 decision making. In this work, we propose a novel 
 attention-based graph convolutional networks model for 
 student's performance prediction. We conduct extensive experiments 
 on a real-world dataset obtained from a large public university.  The 
 experimental results show that our proposed model 
 outperforms 
 state-of-the-art approaches 
 in terms of grade prediction. 
 %
 The proposed  model also shows 
 strong accuracy in identifying students 
 who are at-risk of failing or dropping out 
 so that timely 
 intervention
 and feedback can be provided to the student. 


\end{abstract}

%


\section{Introduction}

Higher educational institutions 
face major 
challenges including timely graduation and retention of enrolled students. The 
National Center for Education Statistics (NCES) 
reports that the six-year graduation rate for 
first-time and full-time undergraduates is around 60\%; the 
retention rate among first-time and full-time 
degree-seeking students is around 80\% \cite{nces}. 
These alarming statistics 
require higher educational institutions to take actions to improve their 
effectiveness and efficiency at educating students. Machine learning 
techniques have been increasingly developed and
applied to educational settings in the hope of improving students' 
learning and increasing students'
success \cite{baker2009state,romero2010handbook,lang2017handbook}. Many systems and applications have 
been proposed; such as course 
recommender systems \cite{elbadrawy2016domain}, academic trajectory and 
degree planning \cite{morsy2019study}, educational early advising systems \cite{hu2017enriching}, and 
knowledge tracing for intelligent tutoring systems 
\cite{yudelson2013individualized, piech2015deep}. Developing methods for
accurate modeling 
and predicting students' performance is the key to 
these systems and applications.

Traditional performance prediction methods can be categorized into two types. The first 
builds a static model, which takes a feature vector as input (such as a 
student's grades in previous courses or student-related features)  and 
outputs   the 
predicted grades. A common approach that 
belongs to this category is linear regression methods
\cite{polyzou2016grade}. 
Students take courses sequentially, i.e., they take some courses at each 
semester; and 
their performance in courses taken in the next semester depends 
on courses taken in previous semesters. Further, 
their knowledge evolves by taking a sequence of 
courses. To capture the temporal dynamics of students' knowledge evolution, 
sequential models have been proposed. A set of representative approaches within 
this category use 
recurrent neural networks (RNN) \cite{kim2018gritnet,hu2019course}.

\begin{figure*}
    \centering
    \begin{subfigure}[b]{.9\textwidth}
        \centering
        \includegraphics[width=\linewidth,height=140pt]{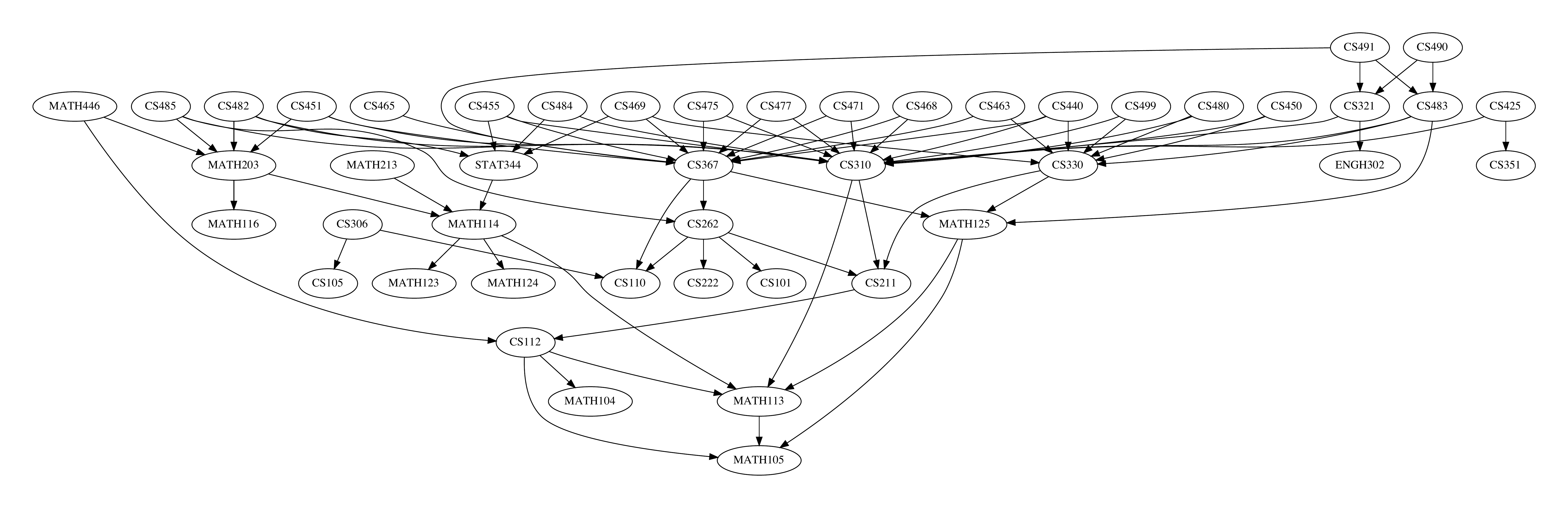}
        \caption{Computer Science} \label{fig:cs_structure}
    \end{subfigure}
    \vfill
    \begin{subfigure}[b]{.9\textwidth}
        \centering
        \includegraphics[width=\linewidth,height=140pt]{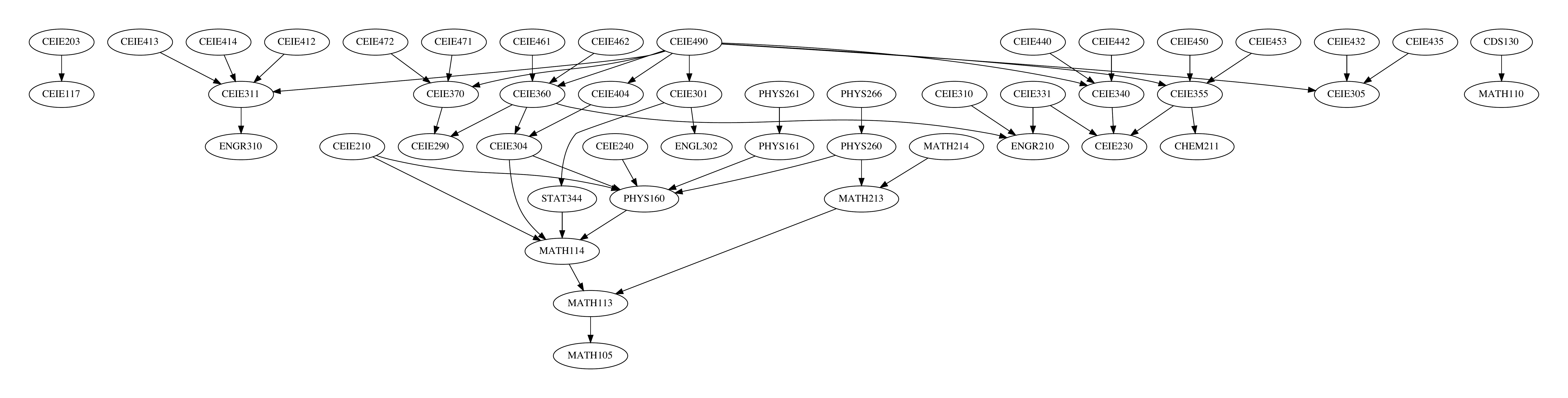}
        \caption{Civil Engineering} \label{fig:ceie_structure}
    \end{subfigure}
    \caption{Course dependence structure in two representative majors.} \label{fig:course_structure}
\end{figure*}

\begin{figure*} [h!]
    \begin{subfigure}[t]{0.3\linewidth}
    \centering
    \includegraphics[width=1.0\textwidth]{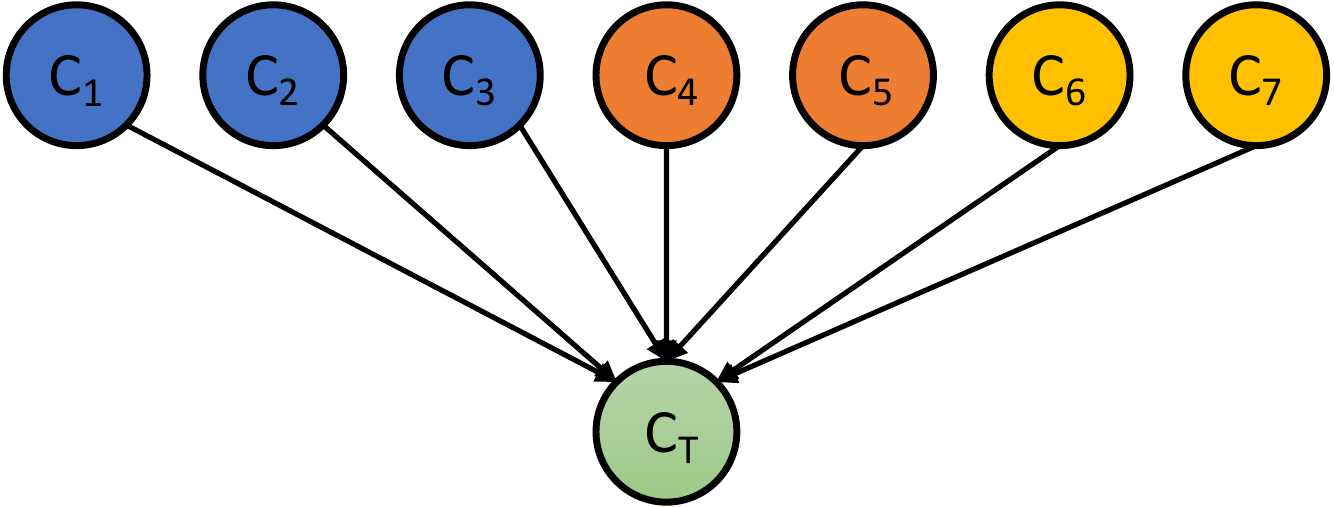}  
    \caption{Static Model} \label{fig:model1}
    \end{subfigure}
    ~
    \begin{subfigure}[t]{0.3\linewidth}
    \includegraphics[width=1.0\textwidth]{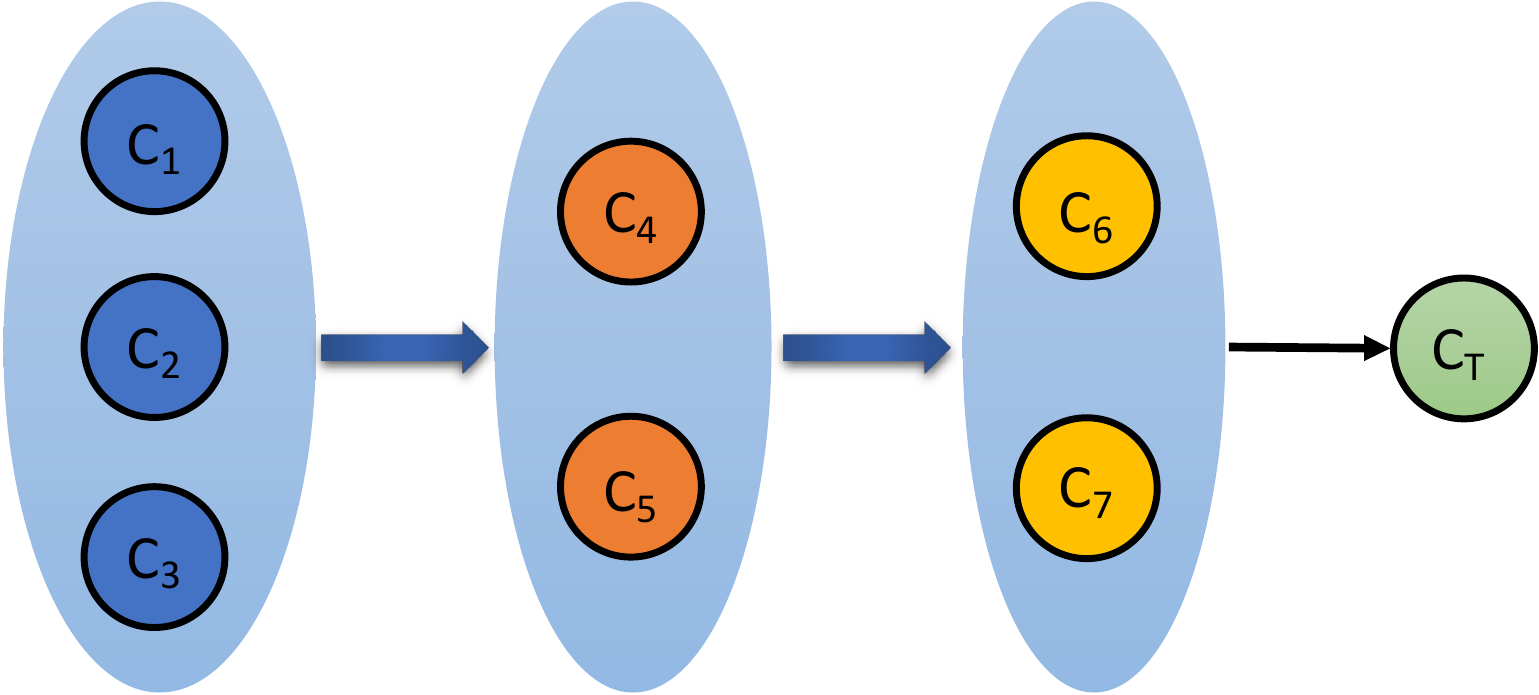}
    \caption{Sequential Model} \label{fig:model2}
    \end{subfigure}
    ~
    \begin{subfigure}[t]{0.3\linewidth}
    \includegraphics[width=1.0\textwidth]{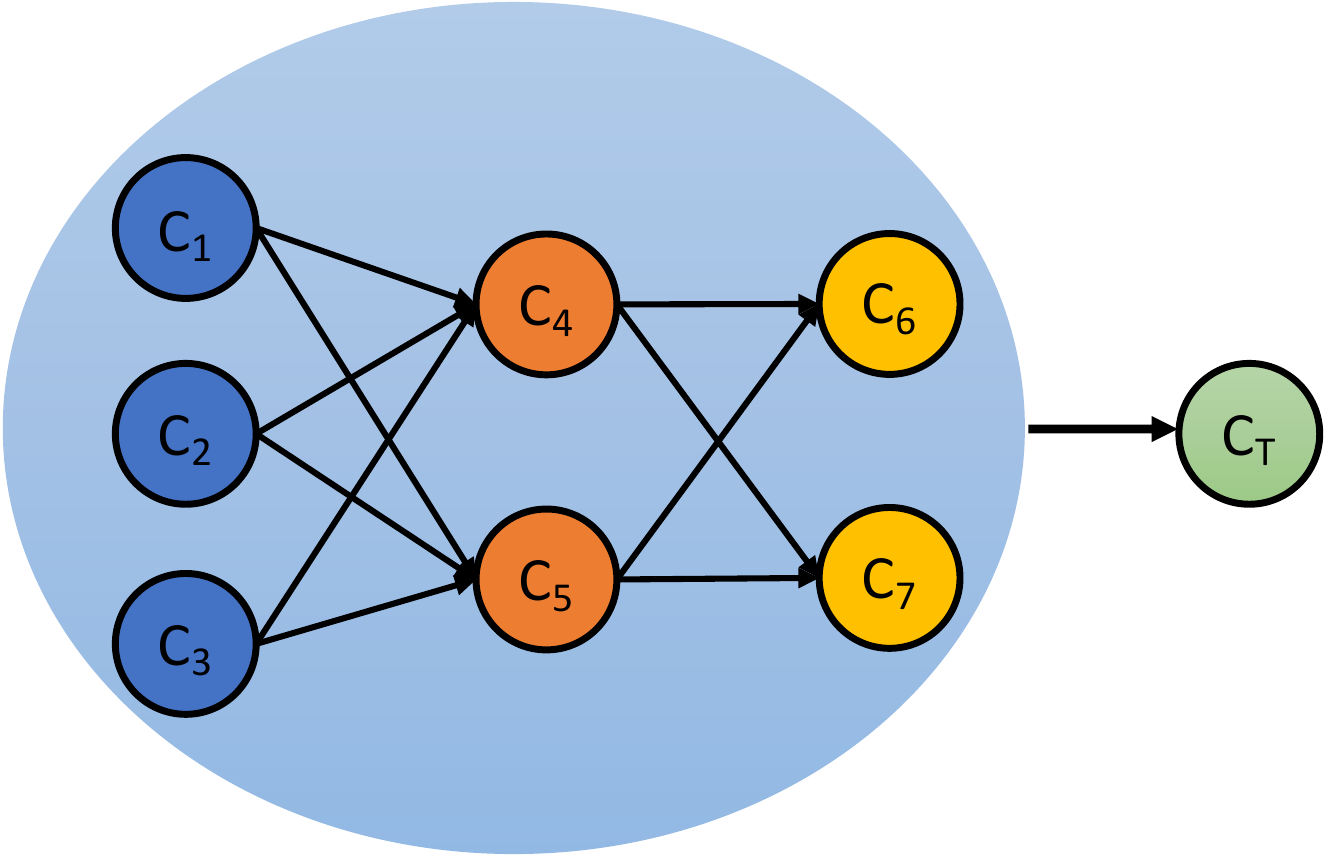}
    \caption{Graph Model} \label{fig:model3}
    \end{subfigure}
    \caption{Comparison of Three Types of Model Architectures. In this example, a student takes course $C_1,C_2,C_3$ in the first semester, course $C_4, C_5$ in the second semester, course $C_6, C_7$ in the third semester before takes the target course $C_T$. } \label{fig:model_comparison}
\end{figure*}

Undergraduate degree programs are designed in a way that knowledge
acquired in prior courses serves as prerequisites
for future courses. The knowledge and skills required to do well 
in a course are acquired in multiple prior courses. The knowledge dependence between courses
exhibit complex graph structure
as shown in Figure \ref{fig:course_structure}. 
Figure \ref{fig:course_structure} shows the prerequisite structures for 
 computer science and civil 
and infrastructure engineering degree programs 
at George Mason University. Each 
node represents a particular
course. An edge pointing from one course to another shows the prerequisite relationship. 
As an example, to do well in the 
data structure course (CS310), students need to acquire
programming skills, object-oriented programming knowledge (CS211) and math (MATH113) which  
come from multiple different courses. The graph in Figure \ref{fig:course_structure} also 
shows hierarchical relationships where a course can depend
on another course which is at a much lower academic level. 
%
%
In addition to the prerequisite structures,
degree programs
are flexible, i.e., 
students can choose to take elective courses based on their interests and do not have to follow a specific ordering 
when taking these courses. 

 

The complexity and flexibility of the degree programs make predicting students' 
performance a challenge task. 
Prior approaches 
usually simplify or ignore 
these complex dependencies. Figure \ref{fig:model_comparison} shows the comparison of 
three types of models. Figure \ref{fig:model1} shows a static model, where 
a student's 
performance is directly dependent on a set of prior courses. Figure \ref{fig:model2} shows a 
sequential model, where students' knowledge evolution is partially modeled.
To overcome the constraints and limitations of the traditional 
models, we propose a model based on graph convolutional networks to 
capture the complex graph-structured knowledge 
evolution exhibited by students' data. Specifically, we propose an 
attention-based graph convolutional network (ACGN) 
model for predicting a student's grade in a future course. Figure \ref{fig:model3} shows the 
graph model, where each course depends on all courses taken in the semester before it so that students' 
knowledge evolution is fully captured. 

When a system is used for decision making e.g., as a support tool 
for advisors to identify students who are at-risk of failing courses 
they will take; it is essential for 
the predictions to be interpretable. This allows the stakeholders 
to trust the decision making systems and make informed decisions. 
We show that our 
attention-based model is able to provide an interpretable 
and useful 
explanation for the predictions. Our model is able to analyze a student's
performance in prior courses and identify 
a collection of important prior courses to explain the student's performance 
in target course.

We performed 
extensive experiments on real-world datasets to evaluate our model 
and compare it with the other two types of 
models aforementioned. The experimental results are 
consistent with our observations that models with architectures 
more close to the degree program have better 
modeling capability and prediction performance. One 
of the important applications for students' performance 
prediction is early warning and advising systems, where at-risk 
students are first identified and timely support is 
provided to improve their academic
success. The experimental results 
show our model's effectiveness at identifying at-risk students.

The key contributions of the paper are summarized as follows:
\begin{itemize}
    \item Flexible graph structured model for students' academic performance prediction. 
    Observing the complex structures of undergraduate degree programs, 
    we propose a graph convolutional network model for students' performance 
    prediction. 
    \item Attention based model for explanation. Providing explanations 
    for a model's predictions makes the model useful for decision 
    making. Our attention-based model can explain the predictions by 
    identifying a set of prior courses important for the predictions.
    \item Identification of at-risk students. While most models achieve good performance 
    at predicting students' performance, they suffer from low accuracy at 
    identifying at-risk students. Our proposed model is able 
    to achieve comparable 
    performance with state-of-the-art models.
\end{itemize}

\section{Related Work}
The need to improve higher 
education services and offerings  has 
attracted research on developing methods for 
predicting students' performance \cite{bakhshinategh2018educational, shahiri2015review}. 
In this section, we review related work on 
students' performance prediction. The related work can 
be classified into three categories: (i) static models, (ii) sequential 
models and (iii) graph models.

\subsection{Static Models}
Static grade prediction models learn a mapping function, where input 
is student-related features
and the output is predicted grade. 
Polyzou et al. \cite{polyzou2016grade} proposed regression models 
specific to courses or students for predicting a 
student's grade in a target course. They found that focusing on a course 
specific subset of the data leads to more accurate predictions. 
Elbadrawy et al. \cite{elbadrawy2015collaborative} introduced 
a personalized multi-regression model for 
predicting students' performance in course activities. Compared to a 
single regression model, this 
model is able to capture personal student differences. To 
understand how students' behavior impacts their 
academic performance, Wang et al. \cite{wang2015smartgpa} collects 
students' behavioral data using smart phone for 
performance prediction. 
Many other classic 
supervised learning approaches 
have been used 
for students' performance prediction including decision trees 
\cite{al2016predicting}, support vector machines and neural 
networks \cite{umair2018predicting}.

Adapted from recommender systems domain, matrix factorization \cite{koren2009matrix} 
approaches are popular for grade prediction. 
%
These factorization approaches
make the 
assumption that a student's 
knowledge/skills and a course's knowledge components can be 
jointly represented with latent vectors (factors) \cite{sweeney2016next}.
Polyzou et al. \cite{polyzou2016grade} 
proposed course-specific matrix factorization models 
for grade prediction that decompose a course-specific 
subset of students' grade data. 
The student course records also exhibit grouping
structures and a domain-aware matrix factorization 
model was developed for the joint course recommendation 
and grade prediction \cite{elbadrawy2016domain}. Ren et al. \cite{ren2017grade} proposed 
matrix factorization model coupled with temporal dynamics for grade prediction. 

 

\subsection{Sequential Models}

%
%

Students take courses sequentially. Their knowledge and skills evolve 
by taking a series of courses. To model the temporal dynamics of students' 
knowledge evolution, sequential models have been proposed. 
Balakrishnan \cite{Balakrishnan:EECS-2013-109} proposed a 
Hidden Markov Model for predicting student 
dropout by modeling students' activities over time in a
Massive Open Online Courses (MOOCs). 
Swamy et al. \cite{swamy2018deep} models student progress on coding assignments in large-scale computer science 
courses using recurrent neural networks.
Kim et al. \cite{kim2018gritnet} 
proposed a bidirectional long short term memory (BLSTM) model for 
the online educational 
setting. Hu et al. proposed course-specific 
markovian models for students' 
grade predictions \cite{hu2018course}. Morsy et al. proposed cumulative knowledge-based 
regression models for next-term grade prediction, which models students' knowledge evolution 
by using a sequential regression model.
Hu et al. \cite{hu2019course} proposed long short term memory models 
for grade prediction in traditional higher education.

\subsection{Graph Neural Networks Models}
Deep learning approaches have found unprecedented success 
in a myriad of applications involving 
regular structured data such as images (grids) and 
text (sequences) \cite{lecun2015deep}. 
Graphs 
are more complex and irregular than grids or 
sequences and recent research efforts involve
designing deep learning models for graph data. Graph
neural networks have been proposed and applied 
to many areas such as 
computer vision for point clouds 
classification \cite{wang2018dynamic}, action 
recognition \cite{yan2018spatial}; recommender 
systems \cite{berg2017graph} and  traffic
prediction \cite{li2017diffusion}. To the best of our 
knowledge, there is no prior 
work on students' performance prediction using graph neural networks.

\section{Methods}

\subsection{Problem Statement}
\begin{figure*} [ht!]
    \centering
    \includegraphics[width=0.8\linewidth]{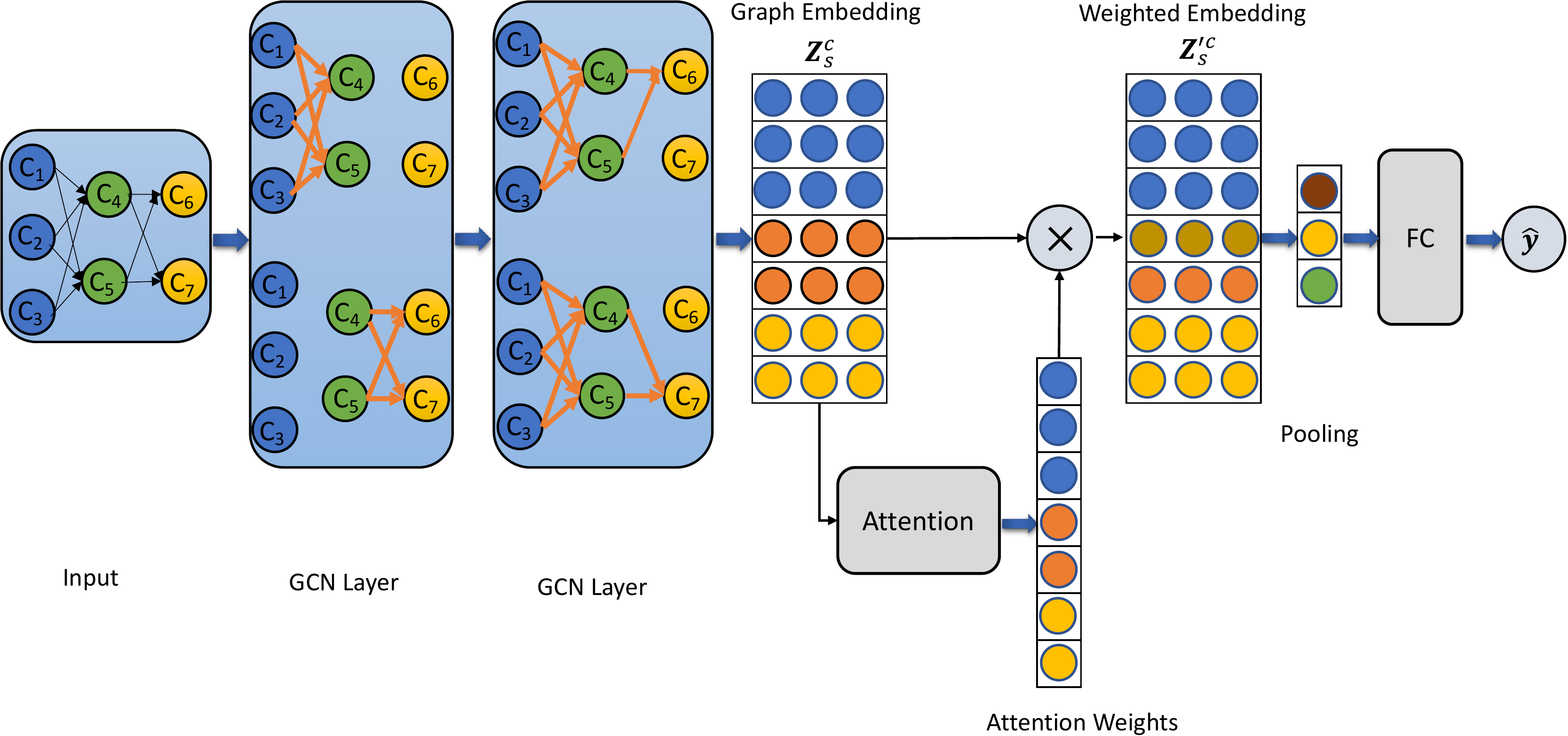}
    \caption{The proposed model.} \label{fig:model_structure}
\end{figure*}



Given a student $s$, the set of courses  taken and grades obtained 
in 
term $t$ are represented by $\mathbb{P}_{s}^t$. 
For 
a sequence of terms $1 \ldots T_s$, we 
denote $\mathbb{P}_s^{1 \sim T_s} =  \mathbb{P}_{s}^{1}, \mathbb{P}_{s}^{2}, \ldots, \mathbb{P}_{s}^{T_s}$
to represent the sequence of courses 
taken and grades obtained by student $s$ in $T_s$ terms. 
%
%
%
%
%
For a target course $c$ taken in the future (next) term, the objective of the proposed 
method is to predict
the grade student $s$ will achieve in course $c$ denoted by $\hat{g}_s^c$. 

%

The proposed models are trained  in a course specific manner i.e., 
for each target course $c$  we learn a unique model. 
%
Due to the flexibility of academic 
degree programs, in each semester 
different courses can be taken; and for each student, the number 
of semesters studied before taking the target 
course will be different.  Therefore, we index the 
length of the sequence with student-specific 
variable $T_s$.

For every target course $c$, a subset of 
frequently taken prior courses are identified from 
all the prior courses taken by students who have already taken the 
target course $c$. These prior courses are denoted as $\mathbb{C}_c$ of 
size $N_c$. 
%
For student $s$, only the prior courses in $\mathbb{C}_c$ are extracted from $\mathbb{P}_s^{1 \sim T_s}$ to 
form a graph which is represented by an 
adjacency matrix $\mathbf{A}_{s}^c \in \{1,0\}^{N_c \times N_c}$ and a 
feature matrix $\mathbf{F}_{s}^c \in \mathbb{R}^{N_c \times D}$, where $D$ represents the number 
of features. 
Take Figure \ref{fig:model3} as an example, the 
student takes courses $c_1, c_2, c_3$ in the
first term, $c_4, c_5$ in the second term and $c_6, c_7$ in the 
third term; we want to predict his/her grade in course $C_T$. Adjacency matrix $\mathbf{A}_{s}^c$ for
this student represents his course taken process. Courses taken in the
current term are fully connected to courses taken in the next term; 1 represents connected, 0 otherwise. 
A row of the feature matrix $\mathbf{F}_{s}^c$ represents the student's grades in 
corresponding prior courses. 



\subsection{Model Description}

Figure \ref{fig:model_structure} shows an 
overview of the 
proposed model.  It is composed of three parts: 1) graph convolutional network, 2) attention layer and 3) a fully 
connected layer.

\subsubsection{Graph Convolutional Network (GCN)}
Convolutional neural networks (CNNs) show superior performance on several applications related to vision \cite{krizhevsky2012imagenet}, speech and text \cite{lecun1995convolutional}. CNNs are powerful because of 
their ability to exploit feature locality at multiple  granularity. 
Graph Convolutional networks have a  similar working 
mechanism but on data with more complex structures, namely, graph.

The input to a 
GCN is an 
adjacency matrix $\mathbf{A}_s^c$ and feature matrix $\mathbf{F}_s^c$, encoding student $s$'s course taking process and grades in prior courses, respectively.  Multiple layers of graph convolutional 
layer are applied on $\mathbf{A}_s^c$ and $\mathbf{F}_s^c$ to learn a 
graph level embedding $\mathbf{Z}_s^c \in \mathbb{R}^{N \times D}$. Each row of $\mathbf{Z}_s^c$ corresponds to a 
node embedding vector. A graph convolutional layer is mathematically described as follows:

\begin{equation}
    \mathbf{H}^{(l+1)} = f(\mathbf{H}^{(l)}, \mathbf{A}) = \sigma (\tilde{\mathbf{D}}^{-\frac{1}{2}} \tilde{\mathbf{A}} \tilde{\mathbf{D}}^{-\frac{1}{2}} \mathbf{H}^{(l)} \mathbf{W}^{(l)})
\end{equation}

where $\tilde{\mathbf{A}} = \mathbf{A} + \mathbf{I}_N$ is the adjacency matrix with self-connections, $\tilde{\mathbf{D}} = \Sigma_j \tilde{\mathbf{A}}_{j}$ is the normalization
matrix, $\mathbf{H}^{(l)}$ is the input and $\mathbf{W}^{(l)}$ is the weight matrix to be learned. 
$\mathbf{H}^{(0)} = \mathbf{F}_s^c$ and $\mathbf{H}^{(L)} = \mathbf{Z}_s^c$; namely, 
the input into the first GCN layer is the feature matrix $\mathbf{F}_s^c$, 
the output from the last GCN layer is the student-specific graph embedding $\mathbf{Z}_s^c$.

A filter in convolutional neural networks aggregates
information from a pixel's neighbors. Similarly, the 
graph convolutional layer aggregates information from a 
node's neighboring nodes and generates a new node embedding vector 
by the following equation
\begin{equation}
\mathbf{h}_{i} = \sigma (\Sigma_j \frac{1}{c_{ij}} \mathbf{h}_{j} \mathbf{W})
\end{equation}
where node $j$ is node $i$'s neighbor.
A higher level of the node embeddings are generated by applying multiple GCN layers. 
Multiple layers of GCN aggregate information from a node's further neighbors. 
As shown in Figure \ref{fig:model_structure}, the first GCN layer aggregates
information from a node's direct neighbors, namely, in our case the courses 
taken in last semester. The second layer collects information from a node's 
second degree neighbors, i.e., the courses taken two semesters ago. The 
final output is the graph embedding 
which entails information from all the courses a student has taken.


\subsubsection{Attention Layer}
The output from GCNs is a graph-level 
embedding matrix, which encodes information about a 
student's knowledge and skills acquired in prior courses. 
The knowledge acquired from different prior courses has different 
importance for the target course. To capture the importance differences 
of the prior courses, we integrate 
attention layer into our model. Attention mechanism allows 
the model to focus on the relevant features or information useful 
for prediction. It works by computing an 
importance score \cite{raffel2015feed}, higher score means the corresponding 
prior course is more important for predicting a student's performance; given by
\begin{equation}
    e_i = MLP(\mathbf{h}_i)
\end{equation}
\begin{equation}
    \alpha_i = \frac{\exp (e_i)}{\Sigma_{k=1}^{N} \exp (e_k)} \label{eq:attention}
\end{equation}
where $MLP$ is a learnable 
function, i.e., multi-layer perceptron, $\alpha_i$ is the attention score 
corresponds to $\mathbf{h}_i$. The output from the attention layer 
is an attention score vector $\bm{{\alpha}}$.

The graph embedding matrix $\mathbf{Z}_s^c$ is weighted by attention scores to form a weighted graph 
embedding matrix $\mathbf{Z}_{ s}^{'c}$ given by
\begin{equation}
\mathbf{Z}_{s}^{'c} = 
  \begin{bmatrix}
    \hdotsfor{1} &     \alpha_1 \mathbf{z}_{s,1}^{c}     & \hdotsfor{1} \\
                 &        \vdots        &              \\
    \hdotsfor{1} &     \alpha_i \mathbf{z}_{s,i}^{c}     & \hdotsfor{1} \\
                 &        \vdots        &              \\
    \hdotsfor{1} &     \alpha_N \mathbf{z}_{s,N}^{c}     & \hdotsfor{1} \\
  \end{bmatrix}
\end{equation}
Finally, the pooling layer coarsens the weighted 
graph embedding matrix into 
a latent vector $\mathbf{v}_s^c$. The latent vector is
passed through a multilayer perceptron; the output from which 
is the predicted grade. 
\begin{equation}
\hat{g}_s^c = f(\mathbf{v}_s^c)
\end{equation}
where $f$ is a multilayer perceptron network.

\section{Experimental Protocol}


\subsection{Dataset Description}

\begin{table}[h!]
    \centering
    \caption{Dataset Statistics} \label{tab:data_stats}
    \begin{adjustbox}{max width=\textwidth}
    \begin{threeparttable}
        \begin{tabular}{c|ccc|ccc}
        \toprule
        \multirow{2}{2em}{Major}
        & \multicolumn{3}{c|}{Fall 2017} & \multicolumn{3}{c}{Spring 2018} \\
        \cline{2-7}
        & \#S & \#C & \#G & \#S & \#C & \#G \\
        \midrule
        CS &  5,042 &  16 &  47,889 &  5,297 &  20 &  52,152 \\
        ECE &  1,992 &  18 &  34,355 &  1,980 &  18 &  34,170 \\
        BIOL &  7,065 &  20 &  52,574 &  6,976 &  20 &  52,672 \\
        PSYC &  5,367 &  20 &  25,207 &  5,368 &  20 &  25,247 \\
        CEIE &  2,222 &  17 &  30,956 &  2,181 &  16 &  30,283 \\
        \hline
        Overall & 21,688 & 91 & 190,981 & 21,802 & 94 & 194,524 \\
        \bottomrule
        \end{tabular}
        \begin{tablenotes}
            \scriptsize
            \item \#S total number of students, \#C number of courses for prediction, \#G total number of grades
        \end{tablenotes}
    \end{threeparttable}
    \end{adjustbox}
\end{table}

The data is collected at George Mason University from
Fall 2009 to Spring 2018. The 
five largest majors are chosen including: 1) Computer Science (CS), 2) Electrical and Computer Engineering (ECE), 3) Biology (BIOL), 4) Psychology (PSYC) 5) Civil Engineering (CEIE). 
The evaluation procedure is
designed in a way to simulate the real-world scenario of
predicting the next-term grades. Specifically, the 
models are trained on the data up to term $T-2$ and 
validated on term $T-1$ and tested on term $T$. The latest two 
terms are chosen as testing terms, i.e. term Fall 2017 and term Spring 2018. For example, to 
evaluate the performance of the models on term 
Fall 2017, the model is trained on data from term Fall 2009 to term Fall 2016,  validated on term Spring 2017 to choose 
the parameters associated with different approaches and finally tested in term Fall 2017. 
The 
statistics of the datasets are listed in Table \ref{tab:data_stats}

\subsection{Evaluation Metrics}
We evaluate the models from 
two perspectives: 1) the accuracy of grade predictions, 2) the models' ability at
detecting at-risk students.

To evaluate the models' accuracy of grade prediction, two evaluation
metrics are used a) mean absolute error (MAE) and 
b) percentage of tick accuracy (PTA).
\begin{equation}
    \text{MAE} = \frac{\sum_{i=1}^N{ | g_i - \hat{g}_i} | }{N}
\end{equation}
where $g_i$ is true grade and $\hat{g}_i$ is predicted grade.

In the grading system, there are 
11 letter grades (A+, A, A-, B+, B, B-, C+, C, C-, D, F) which 
correspond to (4, 4, 3.67, 3.33, 3, 2.67, 2.33, 2, 1.67, 1, 0). A 
tick is  the difference between
two consecutive letter grades. The performance of a model 
is estimated by how many ticks away the 
predicted grade is from the true grade. For example, the 
tick error between B and B is zero, B and B+ is one, B and
A- is two.  
To use PTA for evaluation, we first convert 
the predicted numerical grade to its closest
letter grade and then compute the percentage of 
errors  with 0 tick, within 1 tick,
and within 2 ticks denoted by PTA\textsubscript{0}, PTA\textsubscript{1}, 
and PTA\textsubscript{3}, respectively.


We also evaluate the models' performance of 
identifying 
at-risk students. 
At-risk students are defined as 
those whose grades are 
lower than 2.0 (C, C-, D, F). The predicted grades below 2.0 are treated as 
positives and above 2.0 are treated as negatives. 
The process of detecting at-risk students is similar to grade prediction except that 
the output from the model (the predicted grade) is converted to 1 or 0 based on whether 
the predicted grade is below or above 2.0.
%
As the number 
of at-risk students is low, we use F-1 score as evaluation metric.

\subsection{Comparative Methods}

\subsubsection*{Bias Only (BO)}
Bias only method only takes into account a student's bias, a course's 
bias and global bias\cite{polyzou2016grade}. The predicted grade is as follow
\begin{equation}
    \hat{g}_s^c = b^c + b_s^c + b_{c'}^c
\end{equation}
where $b^c, b_s^c, b_{c'}^c$ are global bias, student bias and course bias, respectively. 

\subsubsection*{Course Specific Matrix Factorization (CSMF)}


The key  assumption underlying this model is that
students and courses can be jointly represented by 
low-dimensional latent factors.
$N$, $M$ and $D$ is the number of students, courses and latent dimension, 
respectively \cite{polyzou2016grade}. To predict a student's grade in a course, we have: 
\begin{equation}
    \hat{g}_s^c = b^c + b_s^c + b_{c'}^c + < \mathbf{u}_s^c, \mathbf{v}_{c'}^c >
\end{equation}
where $b^c$ is global bias, $b_s^c$ is student bias term, $b_{c'}^c$ is course bias term; 
$\mathbf{u}_s^c$ is student $s$'s latent vector, $\mathbf{v}_{c'}^c$ is course $c$'s latent vector.

\subsubsection*{Course Specific Regression (CSR)}
Course specific regression (CSR) \cite{polyzou2016grade} is  a 
linear regression model.
The input into this model is a vector $\mathbf{x}_s^c$ representing a
student's grades in 
prior courses. A course specific subset of prior courses included in 
$\mathbb{P}_s^{1 \sim T_s}$ are 
flattened to form the vector $\mathbf{x}_s^c$.
The predicted grade is
\begin{equation}
    \hat{g}_s^c = w_0^c + \mathbf{x}_s^{c} \mathbf{w}^c
\end{equation}
where $w_0^c$ is bias term and  $\mathbf{w}^c$ are  weight vectors 
 to be learned.

\subsubsection*{Multilayer Perceptron (MLP)}
Multilayer Perceptron is a generalized version of
CSR. CSR model is a linear model, which is not able to 
capture non-linear and complex patterns in 
students' grades data. Therefore, multilayer perceptron 
has been proposed by \cite{hu2019course} for grade prediction. Similar to 
CSR, the input $\mathbf{x}_s^c$ is a student's grades in prior courses. 
\begin{equation}
    \hat{g}_s^c = f(\mathbf{x}_s^c)
\end{equation}
where $f$ is the model to be learned.

\subsubsection*{Long Short Term Memory (LSTM)}
\begin{figure} [h!]
    \centering
    \includegraphics[width=0.9\columnwidth]{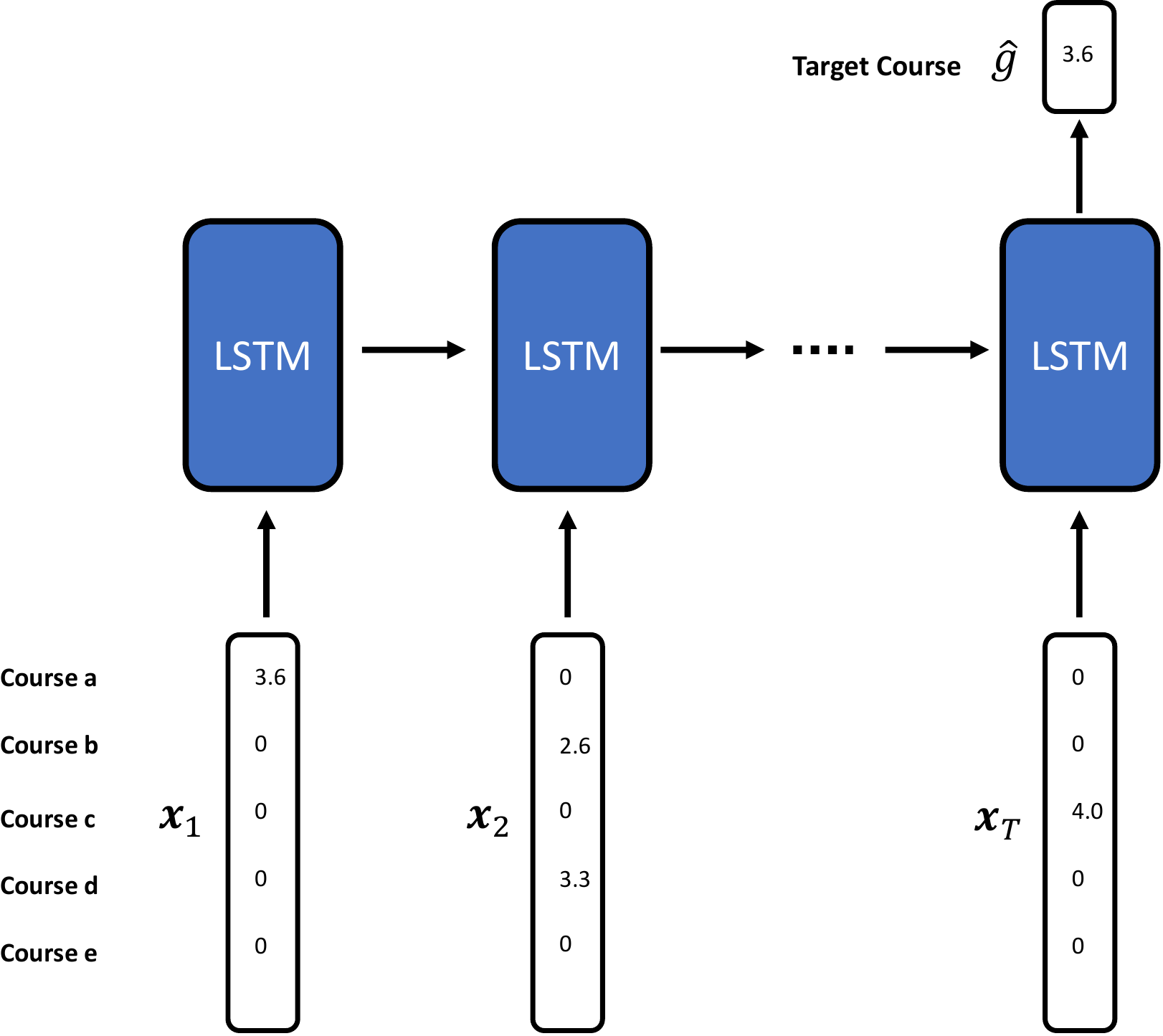}
    \caption{LSTM for grade prediction} \label{fig:rnn}
\end{figure}
Long short term memory (LSTM) is an extension 
of recurrent neural networks (RNN) 
for modeling sequential data. The assumption of using
LSTM for students' performance prediction is that 
students knowledge and skills are evolving by taking courses 
in each semester. To capture the temporal
dynamics of students' knowledge evolution, LSTMs have 
been proposed in \cite{hu2019course}. The input $\mathbf{x}_{s,t}^c$ at time step $t$ 
is a student's grades in courses 
at semester $t$. Many to one architecture is 
utilized and 
the output from the last step of LSTM is fed
into a fully connected network; the output from which is
the predicted grade. The model architecture is 
shown in Figure \ref{fig:rnn},
where the courses $a,b,c,d,e$ are prior courses, $\mathbf{x}_{s,t}^c$ encodes 
the student's grades in courses 
at time $t$ and 
the output $\hat{g}$ is the predicted grade.

\subsection{Implementation}
Our method is implemented in 
Pytorch \cite{paszke2017automatic}. For model optimization
we use Adam \cite{kingma2014adam}. To avoid model
overfitting, we used $l_2$ norm regularization (with coefficient 0.001) and dropout 
(dropout rate 0.05) \cite{srivastava2014dropout}. The number of dimensions 
for the graph embedding is chosen from a list of (8, 12, 16, 20, 32, 64).

\section{Experimental Results}

\subsection{Grade Prediction}
\begin{table*}[h!]
    \centering
    \caption{Comparative Performance of Different Models by MAE. ($\downarrow$ is better)} \label{tab:rslt_mae}
    \begin{adjustbox}{max width=\textwidth}
        \begin{tabular}{c|ccccc|cccccc}
        \toprule
        \multirow{2}{3em}{Method} & \multicolumn{5}{c|}{Fall 2017} & \multicolumn{5}{c}{Spring 2018} \\ \cline{2-11}
         &    CS &   ECE &  BIOL &  PSYC &  CEIE &    CS &   ECE &  BIOL &  PSYC &  CEIE \\
        \midrule
            BO & 0.684 & 0.570 & 0.705 & 0.556 & 0.616 & 0.727 & 0.674 & 0.628 & 0.552 & 0.605 \\
            CSMF & 0.594 & 0.476 & 0.550 & 0.517 & 0.479 & 0.647 & 0.539 & 0.499 & 0.492 & 0.491 \\
            CSR & 0.607 & 0.444 & 0.551 & 0.440 & 0.441 & 0.628 & 0.493 & 0.463 & 0.439 & 0.444 \\
            MLP & 0.585 & 0.390 & 0.515 & 0.407 & 0.413 & 0.590 & 0.436 & 0.417 & 0.413 & 0.369 \\
            LSTM & 0.582 & 0.365 & 0.532 & 0.380 & \bf{0.309} & 0.590 & 0.370 & 0.435 & 0.356 & \bf{0.251} \\
            AGCN & \bf{0.540} & \bf{0.335} & \bf{0.459} & \bf{0.309} & 0.336 & \bf{0.543} & \bf{0.366} & \bf{0.379} & \bf{0.316} & 0.258 \\
        \bottomrule
\end{tabular}
    \end{adjustbox}
\end{table*}

\begin{table*}[h!]
    \centering
    \caption{Comparative Performance of Different Models by Percentage of Tick Accuracy ($\uparrow$ is better)} \label{tab:rslt_tick}
    \begin{adjustbox}{max width=\textwidth}
        \begin{tabular}{c|c|ccccc|ccccc}
        \toprule
        & \multirow{2}{3em}{} & \multicolumn{5}{c|}{Fall 2017} & \multicolumn{5}{c}{Spring 2018} \\\cline{2-12}
        &  Method &  CS &   ECE &  BIOL &  PSYC &  CEIE &    CS &   ECE &  BIOL &  PSYC &  CEIE \\
        \midrule
        \multirow{6}{3em}{\textbf{$\text{PTA}_0$}}
        & BO & 16.76 & 20.75 & 14.40 & 15.52 & 14.90 & 16.07 & 12.11 & 15.42 & 13.65 & 19.79 \\
        & CSMF & 20.00 & 23.58 & 22.40 & 23.10 & 28.85 & 22.31 & 17.37 & 23.35 & 28.41 & 28.65 \\
        & CSR & 24.26 & 33.96 & 27.60 & 38.97 & 40.87 & 26.29 & 28.42 & 34.14 & 41.33 & 35.42 \\
        & MLP & 26.32 & 39.62 & 31.00 & 41.72 & 41.35 & 27.76 & 33.68 & 41.41 & 43.17 & 42.19 \\
        & LSTM & 27.21 & \bf{42.92} & 37.40 & 48.62 & 49.52 & 30.54 & \bf{54.74} & 42.73 & 49.82 & \bf{57.29} \\
        & AGCN & \bf{30.00} & 41.51 & \bf{38.80} & \bf{56.21} & \bf{50.00} & \bf{36.52} & 39.47 & \bf{44.49} & \bf{50.55} & 56.77 \\
        \midrule
        \multirow{6}{3em}{\textbf{$\text{PTA}_1$}} 
        & BO & 44.71 & 49.06 & 43.20 & 57.59 & 48.56 & 44.09 & 37.37 & 46.70 & 57.20 & 50.00 \\
        & CSMF & 55.15 & 62.26 & 60.00 & 63.10 & 62.98 & 52.72 & 54.74 & 63.66 & 59.04 & 61.98 \\
        & CSR & 55.29 & 66.04 & 59.40 & 66.21 & 71.63 & 57.37 & 63.68 & 66.30 & 65.31 & 69.27 \\
        & MLP & 56.91 & 69.81 & 62.80 & 69.66 & 74.52 & 60.03 & 68.42 & 68.28 & 69.37 & 76.04 \\
        & LSTM & 58.24 & 73.11 & 61.40 & 73.79 & 79.33 & 59.10 & 72.11 & 72.03 & 75.65 & 82.81 \\
        & AGCN & \bf{62.21} & \bf{75.47} & \bf{70.00} & \bf{77.93} & \bf{79.81} & \bf{63.61} & \bf{77.89} & \bf{74.89} & \bf{77.86} & \bf{84.90} \\
        \midrule
        \multirow{6}{3em}{\textbf{$\text{PTA}_2$}}
        & BO & 72.94 & 81.13 & 72.40 & 84.83 & 81.25 & 73.97 & 74.21 & 77.75 & 87.45 & 79.17 \\
        & CSMF & 80.00 & 86.79 & 83.60 & 83.45 & 87.50 & 75.30 & 84.21 & 84.36 & 85.24 & 84.38 \\
        & CSR & 76.76 & 86.32 & 80.80 & 83.45 & 84.62 & 77.03 & 82.63 & 84.58 & 82.66 & 86.46 \\
        & MLP & 79.85 & 89.62 & 82.80 & 85.86 & 86.54 & 79.42 & 86.32 & 86.34 & 84.13 & 90.62 \\
        & LSTM & 77.35 & 86.79 & 79.20 & 84.83 & 90.87 & 77.69 & 83.16 & 84.58 & \bf{89.67} & 91.67 \\
        & AGCN & \bf{81.47} & \bf{92.45} & \bf{85.60} & \bf{88.62} & \bf{91.83} & \bf{80.21} &\bf{88.95} & \bf{87.67} & 88.93 & \bf{93.23} \\
        \bottomrule
\end{tabular}
    \end{adjustbox}
\end{table*}

Table \ref{tab:rslt_mae} reports the performance 
of ACGN and comparative approaches for 
the task of next-term grade prediction for the 
Fall 2017 and Spring 2018 semesters using the MAE metric. 
The proposed ACGN 
model achieves the best performance in most cases except the Civil Engineering (CEIE) major. The 
CEIE major has relatively simpler knowledge dependence structure as 
shown in Figure \ref{fig:ceie_structure}. A majority of 
higher level courses, such as 300 and 400 level courses for the 
CEIE 
major have shallow knowledge dependence. While for 
CS major, the higher level courses have deeper knowledge dependence or longer pre-requisite chains. 

Another observation is that models which 
are able to capture the complex
knowledge dependence more have better performance. The static models (BO, CSMF, CSR, MLP) are outperformed by 
sequential model (LSTM) in most cases, on average by 9.2\%; the sequential model is outperformed 
by graph model (AGCN), besides CEIE major, on average by 7.0\%. The experimental results 
are consistent with our assumption that the 
knowledge dependence in the
undergraduate degree programs is complexly 
networked structures and a graph model is well-suited 
at capturing the underlying dynamics. 

Table \ref{tab:rslt_tick} shows the comparative performance using the 
percentage of 
tick error  accuracy.  In contrast to MAE, the PTA metric can 
provide a fine-grained view of the errors made 
by different methods. 
From Table \ref{tab:rslt_tick} we observe
that the performance gap 
between models at \textbf{$\text{PTA}_0$} is larger 
than at \textbf{$\text{PTA}_2$}. For example, for CS majors in 
Fall 2017, the gap between the best performing model AGCN and the 
worst performing model BO at \textbf{$\text{PTA}_0$} is 13.24\%, which is 
larger than 8.53\% at \textbf{$\text{PTA}_2$}.

\begin{table*} [h!]
    \centering
    \caption{Predictive Performance at Identifying At-risk Students, F-1 Score ($\uparrow$ is better)} \label{tab:at_risk}
    \begin{adjustbox}{max width=\textwidth}
    \begin{threeparttable}
        \begin{tabular}{c|ccccc|ccccc}
        \toprule
        \multirow{2}{3em}{Method} & \multicolumn{5}{c|}{Fall 2017} & \multicolumn{5}{c}{Spring 2018} \\\cline{2-11}
        &    CS &   ECE &  BIOL &  PSYC &  CEIE &    CS &   ECE &  BIOL &  PSYC &  CEIE \\
        \midrule
        BO & 0.092 & 0.000 & 0.116 & 0.000 & 0.000 & 0.085 & 0.000 & 0.194 & 0.000 & 0.000 \\
        CSMF & 0.385 & 0.415 & 0.585 & 0.154 & 0.429 & 0.349 & 0.291 & 0.620 & 0.364 & 0.526 \\
        CSR & 0.398 & 0.514 & 0.649 & \bf{0.438} & 0.490 & 0.500 & 0.543 & 0.623 & 0.429 & 0.450 \\
        MLP & 0.383 & 0.426 & 0.630 & \bf{0.438} & 0.500 & 0.534 & 0.472 & 0.676 & 0.400 & 0.605 \\
        LSTM & 0.492 & \bf{0.533} & 0.553 & 0.276 & \bf{0.702} & 0.584 & \bf{0.650} & 0.638 & 0.400 & \bf{0.681} \\
        AGCN & \bf{0.516} & 0.500 & \bf{0.660} & \bf{0.438} & 0.615 & \bf{0.594} & 0.571 & \bf{0.685} & \bf{0.483} & 0.550 \\
        \bottomrule
        \end{tabular}
    \begin{tablenotes}
        \small
        \item The percentage of at-risk students for each major in Fall 2017 is CS (23.7\%), ECE (18.9\%), BIOL (25.8\%), PSYC (8.3\%), CEIE (15.9\%);
        In Spring 2018, it is CS (23.7\%), ECE (24.7\%), BIOL (18.1\%), PSYC (6.6\%), CEIE (14.1\%).
    \end{tablenotes}
    \end{threeparttable}
    \end{adjustbox}
\end{table*}

\subsection{Detecting At-risk Students}
Detecting at-risk students early is a
fundamental task for early warning and advising 
systems. We evaluate the models' performance at
detecting at-risk students. Table \ref{tab:at_risk} shows
the experimental results evaluated by F-1 score. The percentages of at-risk students in 
different majors are presented at the 
table footnote. The 
PSYC major has the lowest percentage 
of at-risk students. The experimental results show that 
LSTM and AGCN achieve the best performance at 
detecting at-risk students. BO performs worst at the 
detection of at-risk students. BO 
only captures the average performance of a 
student and a course, which is biased by other students and courses' performance and the
average performance of other students and courses is usually higher 
than 2.0 (the threshold of defining at-risk students).

\subsection{Interpretation with Attention}
\begin{table*}[h!]
\centering
\caption{Case Studies By Attention Score} \label{tab:case_studies}
\begin{adjustbox}{max width=\textwidth}
\begin{threeparttable}
\begin{tabular}{|c|c|c|c|c|c|}
\hline
Target Course             & True Grade           & Predicted Grade       & Prior Courses & Grades & Attention Score \\
\hline
\multirow{4}{*}{CS-310}   & \multirow{4}{*}{F} & \multirow{4}{*}{C-} & MATH-213      & N      & 0.33            \\
                          &                      &                       & MATH-125      & N      & 0.33            \\
                          &                      &                       & CS-262        & N      & 0.33            \\
                          &                      &                       & \bf{CS-211}        & C    & 0.01            \\
\hline
\multirow{3}{*}{CS-310}   & \multirow{3}{*}{D}   & \multirow{3}{*}{D}    & MATH-213      & F    & 0.913           \\
                          &                      &                       & MATH-114      & C      & 0.072           \\
                          &                      &                       & \bf{CS-211}        & N      & 0.015           \\
\hline
\multirow{2}{*}{BIOL-311} & \multirow{2}{*}{F} & \multirow{2}{*}{C}    & \bf{BIOL-213}      & C+  & 0.5315          \\
                          &                      &                       & \bf{BIOL-214}      & C+   & 0.4685          \\
\hline
BIOL-452                  & D                    & C                     & \bf{CHEM-211}      & C+   & 0.5271          \\
                          &                      &                       & BIOL-214      & B      & 0.2784          \\
                          &                      &                       & \bf{BIOL-213}      & C      & 0.1945          \\
\hline
\end{tabular}
\begin{tablenotes}
    \scriptsize
    \item N means that the student did not take the course. Courses in bold mean they are in prerequisites chain.
\end{tablenotes}
\end{threeparttable}
\end{adjustbox}
\end{table*}

Machine learning models have achieved impressive performance in many tasks. However, 
most of them remain black boxes and there are concerns about their transparency. 
A model's capability to provide explanations for its predictions can increase its 
transparency. For decision making, understanding the reasons behind predictions 
can help decision makers make informed decisions. Grade prediction models 
serve as an assistant tool for advisors to make decisions on whether to 
intervene on a student or not. When the model predicts that a student is at-risk 
of failing a course, knowing which prior courses results in the prediction 
can also help advisors provide personalized feedback to students.

Attention mechanism works by letting the model focus on important information 
for prediction. In our proposed model, the design of the attention layer lets 
the model focus on important prior courses. The output from the attention layer is
a vector of scores representing the importance of the prior courses computed by 
Equation \ref{eq:attention}.
In this section, we show by case studies how the attention scores from the 
attention layer explain the model's predictions, especially, why the model predicts that a 
student is at-risk of failing a target course.

Table \ref{tab:case_studies} shows four case studies. We keep the most important prior courses
identified by attention score.
For the first case study, the target course is CS-310, the student's true grade 
in the target course is F and the predicted grade is C-. The most important four 
courses identified by attention layer is MATH-212, MATH-125, CS-262, CS-211. The 
reason for predicting this student as at-risk is that the student did not take MATH-212, MATH-125, CS-262, 
therefore lacks the
necessary knowledge to do well in the target course. 
In the second case, the student's true grade in CS-310 is D, the predicted grade is D. 
The three most important courses are MATH-213, MATH-114, CS-211. The reason 
for predicting this student as failing the target course is that he failed MATH-213 
and did not do well in MATH-114 and did not take CS-211, which is the prerequisite of the target course. 
In the third case, the student's true grade in the target course is F and the predicted grade is C.
The two most important prior courses identified are BIOL-213 and BIOL-214, both are in 
prerequisite chain of the target course and the student did not do well in them.
The fourth case shows that the student failed the target course BIOL-452 and the predicted grade is C.
The three most influential prior courses are CHEM-211, BIOL-214, BIOL-213. Courses CHEM-211 and 
BIOL-213 are in prerequisite chain and the student did not perform well in them.

From the case studies, we can see that the attention layer 
identifies missing knowledge components for a target course, arising due 
to two reasons: 1) the student did not take some important prior courses, 
2) the student did not do well in the corresponding prior courses.

\subsection{Sensitivity Analysis}
\begin{figure} [th!]
    \centering
    \includegraphics[width=\linewidth]{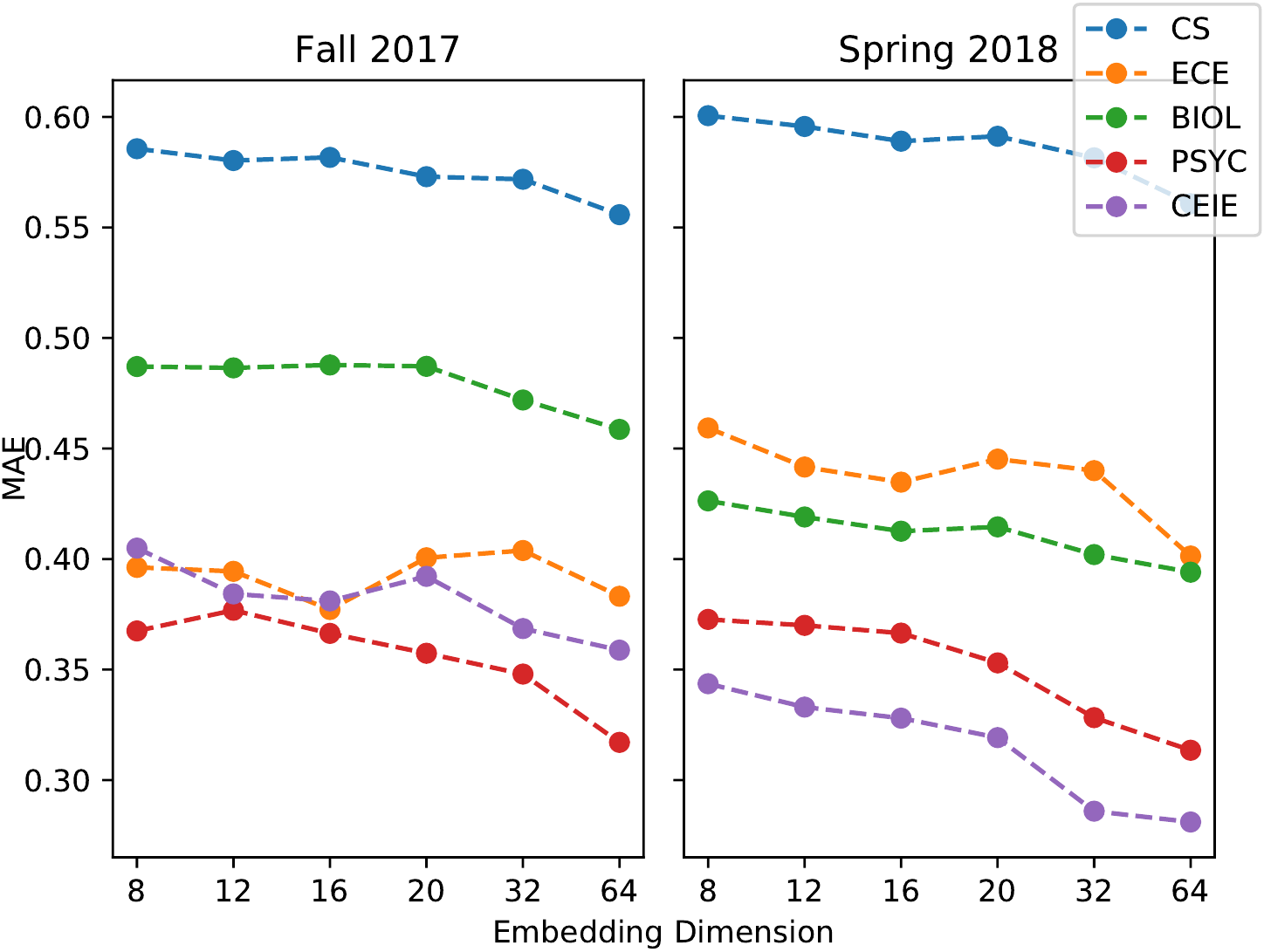}
    \caption{Sensitivity analysis on embedding dimension.} \label{fig:sens_analysis}
\end{figure}

In this section, we evaluate the sensitivity 
of the model's performance with respect 
to the dimension of the graph embedding. In Figure \ref{fig:sens_analysis}, the 
x-axis is the embedding dimension and y-axis is MAE for Fall 2017 and Spring 2018 datasets. 
From Figure \ref{fig:sens_analysis}, we can see that the 
model's performance varies with the dimension size. Overall, its performance is quite stable across the 
different majors.

\section{Conclusions}
Students' performance prediction is a fundamental task in educational data mining. Predicting students' performance
in undergraduate degree programs is a challenging task due to several reasons. First of all, 
undergraduate degree programs exhibit complex knowledge 
dependence structures. Secondly, undergraduate degree programs are flexible 
which means students can take courses without 
following specific order
and they can choose to take whatever electives they are interested in. Traditional approaches like
static and sequential models are not able to fully capture the complexity and flexibility of students' data.

In this work, we proposed a novel attention-based graph convolutional networks for students' performance 
prediction. The model is able to capture the relational 
structure underlying students' course records data. We performed
extensive experiments to evaluate the proposed model on real-world datasets.
The model is evaluated in several aspects: 1)  grade prediction accuracy and 2)  ability to 
detect at-risk students. The experimental results show that our model 
outperformed state-of-the-art approaches in terms of both grade prediction accuracy and at-risk students detection. 
Finally, the attention layer provides explanations 
for the model's prediction, which is essential for decision making.

\section{Acknowledgements}
This work was supported by the National Science Foundation grant \#1447489. The computational resources was provided by ARGO, a research computing cluster provided by the Office of Research Computing at George Mason University, VA. (URL:http://orc.gmu.edu)

%
\bibliographystyle{abbrv}
\bibliography{refs}  
%
%

\balancecolumns
\end{document}